\documentclass[11pt]{article} 
\usepackage[top=1in, bottom=1in, left=0.75in, right=.75in]{geometry}
\usepackage{times}
\usepackage{authblk}  

\usepackage{amsmath,amsfonts,bm}

\def\eqref#1{equation~\ref{#1}}

\def\1{\bm{1}}

\DeclareMathAlphabet{\mathsfit}{\encodingdefault}{\sfdefault}{m}{sl}
\SetMathAlphabet{\mathsfit}{bold}{\encodingdefault}{\sfdefault}{bx}{n}

\usepackage{hyperref}
\usepackage{url}
\usepackage{graphicx}
\usepackage{booktabs}
\usepackage{natbib}
\usepackage{fancyvrb}
\usepackage{xcolor}

\title{Question Answering under Temporal Conflict: Evaluating and Organizing Evolving Knowledge with LLMs}

\author{\Large Atahan Özer}
\author{\Large Çağatay Yıldız}

\affil{~\large University of Tübingen}
\date{}  

\begin{document}

\maketitle

\begin{abstract}
Large language models (LLMs) exhibit remarkable capabilities in question answering and reasoning thanks to their extensive parametric memory.  However, their knowledge is inherently limited by the scope of their pre-training data, while real-world information evolves continuously. Updating this knowledge typically requires costly and brittle re-training, or in-context learning (ICL), which becomes impractical at scale given the volume and volatility of modern information. Motivated by these limitations, we investigate how LLMs perform when exposed to \emph{temporal text corpora}, or documents that reflect evolving knowledge over time, such as sports biographies where facts like a player's ``current team'' change year by year. To this end, we introduce two new benchmarks: \textit{Temporal Wiki}, which captures factual drift across historical Wikipedia snapshots, and \textit{Unified Clark}, which aggregates timestamped news articles to simulate real-world information accumulation. Our analysis reveals that LLMs often struggle to reconcile conflicting or outdated facts and can be misled when multiple versions of a fact appear in context. To address these issues, we propose a lightweight, agentic framework that incrementally builds a structured, external memory from source documents without requiring re-training. This knowledge organization strategy enables models to retrieve and reason over temporally filtered, relevant information at inference time. Empirically, our method outperforms ICL and RAG baselines across both benchmarks, especially on questions requiring more complex reasoning or integration of conflicting facts.
\end{abstract}

\section{Introduction}

The world evolves continuously, and so does the information that reflects its state. Large language models (LLMs) can encode a snapshot of the world in their parametric memory, but only up to the limits of their training data and cut-off dates. 
When tasks require knowledge beyond a model’s training horizon, two main strategies emerge: in-context learning or re-training. In-context learning has proven effective in temporal settings \citep{temporal_train, temp_train2, temp_train3}, and is widely adopted in state-of-the-art chatbots, e.g., by integrating web search to handle queries involving post-training facts. In contrast, re-training falls under the broader umbrella of continual learning, which explores how LLMs can incrementally absorb new information over time \citep{caca_pre}. However, most existing approaches predate modern LLM architectures. As LLMs continue to scale, both re-training and continual adaptation become increasingly resource-intensive.

While in-context learning (ICL) has been explored across various tasks, such as few-shot classification \citep{brown2020language}, compositional reasoning \citep{wei2022chain}, and tool use \citep{schick2023toolformer}, its application to \emph{temporal contexts} remains relatively underexplored. 
Related to this, recent work on knowledge conflict \citep{stub_conflict, joon_conflict} investigates how LLMs handle discrepancies between (outdated) parametric memory and the external information in their context. These studies show that LLMs can align with external facts when the input is coherent and unambiguous. However, when both supporting and conflicting evidence appear in the same context, models, especially larger ones, tend to fall back on their parametric knowledge. This behavior is particularly concerning in temporal settings, where outdated facts are common and can lead to degraded performance. Furthermore, using ICL to address temporally evolving knowledge introduces known challenges, including performance degradation with longer inputs \citep{long1, relevancy, long_rag}, sensitivity to answer position \citep{lost_in_middle}, and difficulty identifying relevant context \citep{relevancy}. These limitations motivate the need for a deeper investigation of ICL within temporal frameworks.

Although temporal ICL with LLMs is a new field, important foundational work was done before the rise of LLMs. For instance, Templama \cite{templama} investigates how to train language models on temporal facts and introduces one of the first temporal Wikipedia datasets. Other early work explores temporal misalignment, analyzing how the classification accuracy of pretrained or adapted language models changes when the temporal context of input and training data diverge \citep{temp_misalign}. Additionally, research on temporal reasoning in LLMs examines how well models localize events in time and handle time-dependent arithmetic, revealing that temporal logic is a challenge for LLMs \cite{temp_reason1, temp_reason2}. 
However, these studies largely overlook the problem of \emph{continually evolving knowledge}, the focus of our work, which involves not just temporal reasoning but also the accumulation and integration of new facts over time.

Closest to ours are two recent works that investigate temporality, but from different perspectives. The first is Erase \cite{erase}, which develops a mechanism to dynamically build and edit a knowledge base to align an LLM's external knowledge with new information. They introduce the Clark dataset, which we also use in our experiments. The second is the Growover \cite{growover}, which, similar to us, uses Wikipedia to simulate evolving knowledge. They propose a reflection method called the ``retrieval-interactive language model''. We note that these two works do not study the temporal aspect of the task, i.e., how LLMs handle conflicting knowledge or when a question requires a list as the answer (e.g., the teams a basketball player played for).

To this end, we introduce two new benchmarks, \textit{Temporal Wiki} and \textit{Unified Clark}, that simulate the dynamics of evolving and accumulating knowledge in natural language. Our experiments reveal that temporal data poses challenges for LLMs, particularly in handling knowledge drift and resolving memory conflicts. To tackle these issues, we propose a novel computational paradigm that incrementally organizes external knowledge in the documents into structured text. This approach enables models to reason over temporally coherent information without re-training, and yields consistent improvements on tasks requiring adaptation to changing facts.

\section{Methodology}
Next, we describe the motivation of this work, our introduced datasets, details of the experiments, and our agentic framework solving this problem, called \textbf{knowledge organization}.

\subsection{Motivation} 
Pre-LLM temporal question answering primarily focused on self-localization in time or answering time-sensitive questions through supervised training. With the rise of in-context learning (ICL), open-book question answering (particularly retrieval-augmented methods) has received increasing attention. In many real-world scenarios, retrieved documents contain unstructured and temporally varying information. For instance, answering a question like \textit{``Which teams has LeBron James played for?''} requires aggregating evidence from sources spanning different time periods, where multiple documents may reference different teams due to temporal shifts. As we show empirically, such questions whose answers evolve over time pose significant challenges for LLMs. This work investigates how well LLMs handle these temporally dynamic, real-world conditions.

\subsection{Datasets} 
We introduce two new benchmarks inspired by prior work. The first, \textbf{Temporal Wiki}, builds on the Templama dataset \citep{templama} and is designed to study how knowledge in Wikipedia evolves over time and how LLMs interact with this evolving information. The second, \textbf{Unified Clark}, extends the ERASE dataset \citep{erase} to evaluate LLMs’ performance in real-world temporal information retrieval scenarios using news data. In the following subsections, we describe the construction, processing, and intended use of these benchmarks for evaluating temporal understanding in LLMs. The \textbf{Temporal Wiki} dataset is publicly available at \url{https://github.com/atahanoezer/TQA}, while the \textbf{Unified Clark} dataset can be accessed through \citep{erase}.

\subsubsection{Temporal Wiki}
Templama is a dataset composed of time-sensitive \texttt{(Subject, Relation, Object)} triplets extracted from Wikipedia (e.g., \textit{{(LeBron James, play for, Los Angeles Lakers)}}). Its original purpose was to address the limitations of language models regarding outdated knowledge due to training data cut-off dates, and to explore techniques for integrating new temporal information during pretraining. 
In this work, we repurpose Templama by leveraging its direct link to Wikipedia: for each triplet, we collect multiple historical snapshots of the subject’s Wikipedia page across time, enabling us to simulate both knowledge accumulation and temporal drift in question-document pairs derived from these snapshots.

Our data curation pipeline begins by constructing question-document pairs where the document contains the answer to a generated, time-dependent question. We first convert each Templama triplet into a temporal question, for instance, \textit{``What team did LeBron James play for in \textbf{2014}?''}. For every subject, we use the Wikidata API to download all available snapshots of its Wikipedia page. Each question is then paired with the earliest snapshot from the year following the query time (e.g., \textit{January 2015} for a 2014 question), ensuring that the relevant information has likely been added to the page. If a snapshot from the target year is unavailable, we fall back to the closest available version that follows the question’s timestamp.

To ensure data quality, we apply the following post-processing steps:

\begin{itemize}
    \item \textbf{Removing structured knowledge removal:} Wikipedia often includes structured knowledge elements, such as tables and bullet-point lists, that condense information in ways not representative of natural unstructured text. These elements are removed using regular expressions to better emulate real-world information retrieval conditions.
    \item \textbf{Character standardization:} Raw Wikipedia snapshots frequently contain non-human-readable artifacts, such as hyperlinks, markup, and ASCII escape sequences. We remove these elements from the text using regex-based preprocessing to produce clean, readable documents suitable for LLM inputs.
    \item \textbf{Evidence answer check:} 
    Manual inspection revealed that not all paired documents contain an answer to the corresponding question. To mitigate this, we employ a two-step filtering process. First, a rule-based answer checker scans the reference document for surface-level matching with the expected answer. Second, we apply a semantic verification step using GPT-4o to confirm whether the reference indeed supports the answer. 
\end{itemize}

\begin{figure*}[t!]
    \centering
    \includegraphics[width=0.98\linewidth]{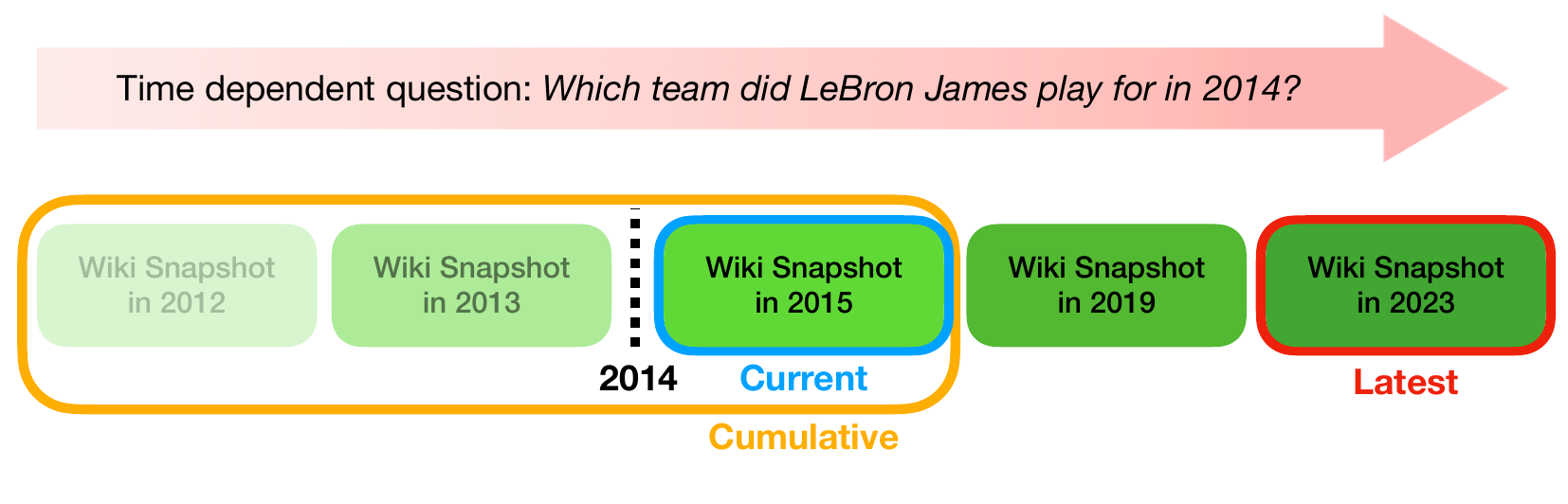}
    \vspace*{-.5cm}
    \caption{An example question from our Temporal Wiki dataset and different contextual evidence. The question dates back to 2014. We illustrate the closest (the snapshot from the year nearest to the question incident), the cumulative (the snapshot that includes all information up to the question year), and the latest snapshot (the most recent snapshot available).}  
    \label{fig:qa_pipeline}
\end{figure*}

\paragraph{Temporal Wiki snapshots}
Next we describe three different ways of forming contextual evidence in our experiments. We define facts $\mathbf{F_i}$ as comprising yearly incident questions $\mathbf{Q_{i,t}}$ and year-dependent reference texts $\mathbf{R_{i,t}}$, where $i \in \{1, \dots, N\}$ and $t \in \{1, \dots, T\}$. We consider three categories of contextual evidence (see Figure~\ref{fig:qa_pipeline} for a visual illustration):
\begin{itemize}
    \item \textbf{Closest snapshot} corresponds to the pair $(\mathbf{R_{i,t}, Q_{i,t}})$, where the reference text aligns temporally with the question. This setup ensures that the correct answer is in the text while previous answers are also likely to be mentioned.
    \item \textbf{Latest snapshot} uses $(\mathbf{R_{i,T}, Q_{i,t}})$, where the most recent available reference text is paired with the question. This setup simulates a more complex knowledge base than the above.
    \item \textbf{Cumulative snapshot} considers $(\mathbf{R_{i,1:t}, Q_{i,t}})$, which aggregates all reference texts from the first to the current year $t$. This setup accounts for the occasions where information is deleted from Wikipedia. Hence, we achieve highest degree of information coverage by this approach while the text gets longer.
\end{itemize}

\subsubsection{Unified Clark}
Wikipedia pages are cumulatively updated over time, meaning that a snapshot from a given year (e.g., LeBron James’ 2015 Wikipedia page) typically contains both the most recent answer as well as all prior answers to a question. In contrast, the Clark dataset is composed of news articles with timestamps, where each article generally reflects only the current answer at the time of publication. As a result, when used in isolation, individual news articles make ICL straightforward for LLMs, since no conflicting information is present.

To introduce a level of complexity comparable to our Temporal Wiki dataset, we concatenate all available news articles for a given entity (along with their timestamps) into a single document. We refer to this extended version as \emph{Unified Clark}. Additionally, because many articles in the original dataset do not explicitly emphasize temporal details, we frame questions to require comprehensive answers that span multiple time periods (e.g., \textit{``What teams did LeBron James play for?''}). Our primary goal in using this dataset is to study memory conflict: scenarios in which multiple, and sometimes conflicting, answers co-occur within the context, presenting a greater challenge for LLMs.



\subsection{Experiment details}

\paragraph{Models}
We conducted our experiments using on-premise models: Llama 3.1 70B \citep{grattafiori2024llama}, Llama 3 8B \citep{grattafiori2024llama}, and Mistral 7B Instruct v2 \citep{mistral2024mistral7bv2}. For evaluation purposes, we additionally used Qwen-2 72B \citep{chu2024qwen2}. The 7B and 8B models were deployed on a single NVIDIA A100 GPU, while the 70B+ models were distributed across three A100 GPUs using 4-bit quantization. All models were served using the Hugging Face Transformers library \citep{huggingface}, with text tokenization handled by each model’s native byte pair encoding (BPE) tokenizer. Quantized inference was enabled via the \texttt{bitsandbytes} library \citep{dettmers2022optimizers}.

\paragraph{Prompting} 
Recent studies have shown that the prompting strategy used during LLM inference plays a critical role, as well as the quality of the model’s parametric memory and the relevance of the provided context \citep{cot, self_refinement, meta_prompting}. In our experiments, we adopt the Meta Prompting approach proposed by \citet{meta_prompting}, which combines a small number of illustrative examples with clear, task-specific instructions. This strategy consistently improves LLM performance across diverse settings. A representative example of our prompting template is provided in the appendix.

\paragraph{RAG Setup} 
We split documents into chunks of 500 characters with a 50-character overlap using the ``Recursive Character Text Splitter'' from the LangChain library. The chunks are encoded into a 512-dimensional embedding space. At inference time, we retrieve the top 12 most relevant chunks by computing cosine similarity between the query embedding and all document embeddings. Note that we obtained these hyperparameters after exhaustive ablation search.

\paragraph{Evaluation} 
Our benchmark requires comparing open-ended model outputs by human-annotated gold answers. 
While traditional string-matching metrics such as Exact Match (EM) and token-level F1 score are widely used, they often fail to capture semantic equivalence \citep{semantic_eval}. One of the most reliable evaluation methods is human evaluation with cross-examination; however, this approach is costly and may introduce additional biases \citep{llm_eval_1}. Fortunately, as demonstrated by recent studies \citep{llm_eval_1,llm_eval2}, state-of-the-art LLMs are capable of achieving human-level performance in evaluation tasks, where evaluation agreement exceeds 80\%, comparable to human evaluators. Motivated by these findings, we adopt LLM-based evaluators for our experiments. To mitigate individual model biases, we use a consensus voting scheme between Llama 3.1 70B and Qwen-2 72B. Further details on the evaluation setup and reliability analysis are provided in the Appendix.

\begin{figure*}[b!]
    \centering
    \includegraphics[width=0.98\linewidth]{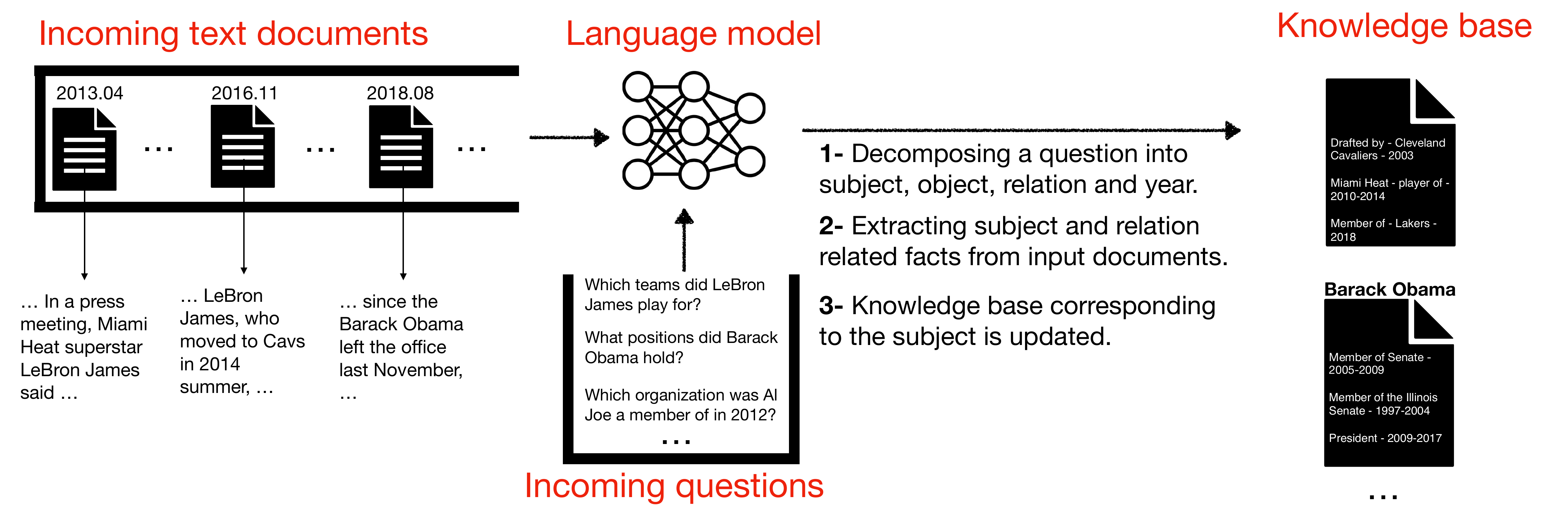}
    \caption{Visual summary of our agentic workflow, where incoming text documents are turned into entries in a structured knowledge base in light of the subjects, objects, and relations extracted from the questions.}  
    \label{fig:ko}
\end{figure*}

\subsection{Our knowledge organization approach}
Finally, we introduce a straightforward but effective agentic solution to the challenge of question answering over temporally evolving knowledge. 
Inspired by how humans acquire and structure knowledge, our method builds a dynamic knowledge base from evidence documents and performs reasoning over this structured knowledge base. This setup encourages precision by ensuring that the model reasons only over relevant, structured knowledge, instead of relying on large, unfiltered context windows. Please see Figure~\ref{fig:ko} for a summary of our approach.

The workflow of our approach proceeds in three stages. First, given an incoming question and an associated evidence document, the agent uses in-context learning to convert the question into a structured quadruple of the form {\texttt{(Subject, Relation, Object, Timestamp)}}. In the second stage, the agent extracts all relevant facts from the document concerning the same subject, again using in-context learning. These extracted facts are then stored as structured entries in the agent’s knowledge base, indexed by subject and relation. In the final stage, when answering the question, the agent queries its knowledge base for all entries matching the subject and relation in the question. It then formulates an answer using only the retrieved entries.

\section{Experiments}
We present our experimental results on the \textit{Temporal Wiki} and \textit{Unified Clark} datasets, evaluating models under four inference schemes: \textbf{zero-shot (ZS)}, \textbf{in-context learning (ICL)}, \textbf{retrieval-augmented generation (RAG)}, and \textbf{knowledge organization (KO)}. In the \textbf{ZS} setting, the model answers each question without access to any supporting documents, relying entirely on its parametric memory. In contrast, the \textbf{ICL}, \textbf{RAG}, and \textbf{KO} settings provide external documents as input, each leveraging contextual information in different ways to improve performance.

\subsection{Temporal Wiki}
We now present our findings on the Temporal Wiki benchmark. To better understand model behavior, we categorize questions into two groups based on whether they are answered correctly in the absence of external evidence.\footnote{The zero-shot accuracies for Llama 3.1 70B, Mistral 7B v2, and Llama 3 8B are 0.62, 0.36, and 0.43, respectively.} Table~\ref{tab:zero_minus} reports the performance when external context is introduced, i.e., the cases where the model's accuracy decreases due to conflicting or distracting information in the input. Conversely, Table~\ref{tab:zero_plus} shows accuracy on questions that were originally answered incorrectly in the zero-shot setting but improved when supporting context was added. To further analyze the impact of information quality, we present ablation studies in Figures~\ref{fig:fact-change-accuracy} and \ref{fig:char-length-accuracy}, focusing on the effects of how many times the answer to a question has changed over time and document length, respectively.

\begin{table}[!ht]
    \centering
    \caption{Accuracy computed on questions whose zero-shot answers are correct. We observe that KO causes the least degradation in accuracy.}
    \vspace*{.5cm}
    \resizebox{\textwidth}{!}{%
   \begin{tabular}{c|c|c|c|c|c|c}
        \toprule
        \textbf{Model} & \multicolumn{2}{c|}{\textbf{ICL}} & \multicolumn{3}{c|}{\textbf{RAG}} & \textbf{Knowledge Org.} \\
        \cmidrule(lr){2-3} \cmidrule(lr){4-6}
        & \textbf{Closest} & \textbf{Latest} & \textbf{Closest} & \textbf{Latest} & \textbf{Cumulative} & \\
        \midrule
        Llama-3.1 70B & 0.83 & 0.76 & \textbf{0.92} & 0.90 & 0.90 & 0.90 \\
        Llama-3 8B & 0.85 & 0.78 & 0.86 & 0.82 & 0.83 & \textbf{0.92} \\
        Mistral 7B v2 & 0.80 & 0.76 & 0.86 & 0.81 & 0.81 & \textbf{0.91} \\
        \bottomrule
    \end{tabular}%
    }
\label{tab:zero_minus}
\end{table}

\begin{table}[!ht]
    \centering
    \caption{Accuracy computed on questions whose zero-shot answers are wrong. The results show that KO utilizes the external input the best when answering the questions, followed by RAG.}
    \vspace{.5cm}
    \label{tab:zero_plus}
    \resizebox{\textwidth}{!}{%
    \begin{tabular}{c|c|c|c|c|c|c}
        \toprule
        \textbf{Model} & \multicolumn{2}{c|}{\textbf{ICL}} & \multicolumn{3}{c|}{\textbf{RAG}} & \textbf{Knowledge Org.} \\
        \cmidrule(lr){2-3} \cmidrule(lr){4-6}
        & \textbf{Closest} & \textbf{Latest} & \textbf{Closest} & \textbf{Latest} & \textbf{Cumulative} & \\
        \midrule
        Llama-3.1 70B & 0.65 & 0.62 & 0.78 & 0.77 & 0.74 & \textbf{0.81} \\
        Llama-3 8B & 0.69 & 0.61 & 0.73 & 0.63 & 0.66 & \textbf{0.83} \\
        Mistral 7B v2 & 0.70 & 0.60 & 0.72 & 0.68 & 0.68 & \textbf{0.82} \\
        \bottomrule
    \end{tabular}%
    }
\end{table}


\subsubsection{General observations}
We begin with observations that hold consistently across both question categories:
\begin{itemize}
    \item Our \textbf{knowledge organization (KO)} approach consistently outperforms both ICL and RAG across all models, except for the RAG setup using Llama 3.1 70B. These results suggest that additional computation to distill documents into structured, dense knowledge is beneficial for downstream reasoning.
    \item \textbf{RAG consistently outperforms ICL.} This aligns with expectations since retrieval serves as a filtering mechanism that limits the input to question-relevant document chunks. This reduces the risk of injecting distracting or irrelevant information into the context window.
    \item Both ICL and RAG benefit from \textbf{shorter contextual inputs.} Across all models and question types, using the \textit{closest} snapshot (i.e., temporally nearest Wikipedia snapshot to the query) yields better results than using the \textit{latest} snapshot. Combined with the previous observation, this suggests that excessive temporal context may introduce noise and confuse the model.
    \item \textbf{Larger models tend to perform better.} As expected, larger LLMs generally achieve higher accuracy, thanks to their greater capacity for reasoning and retention \citep{parametric_mem}.
\end{itemize}

\subsubsection{Setup-specific observations}

\begin{figure}[!t]
    \centering
    \begin{minipage}[t]{0.48\textwidth}
        \centering
        \includegraphics[width=\linewidth]{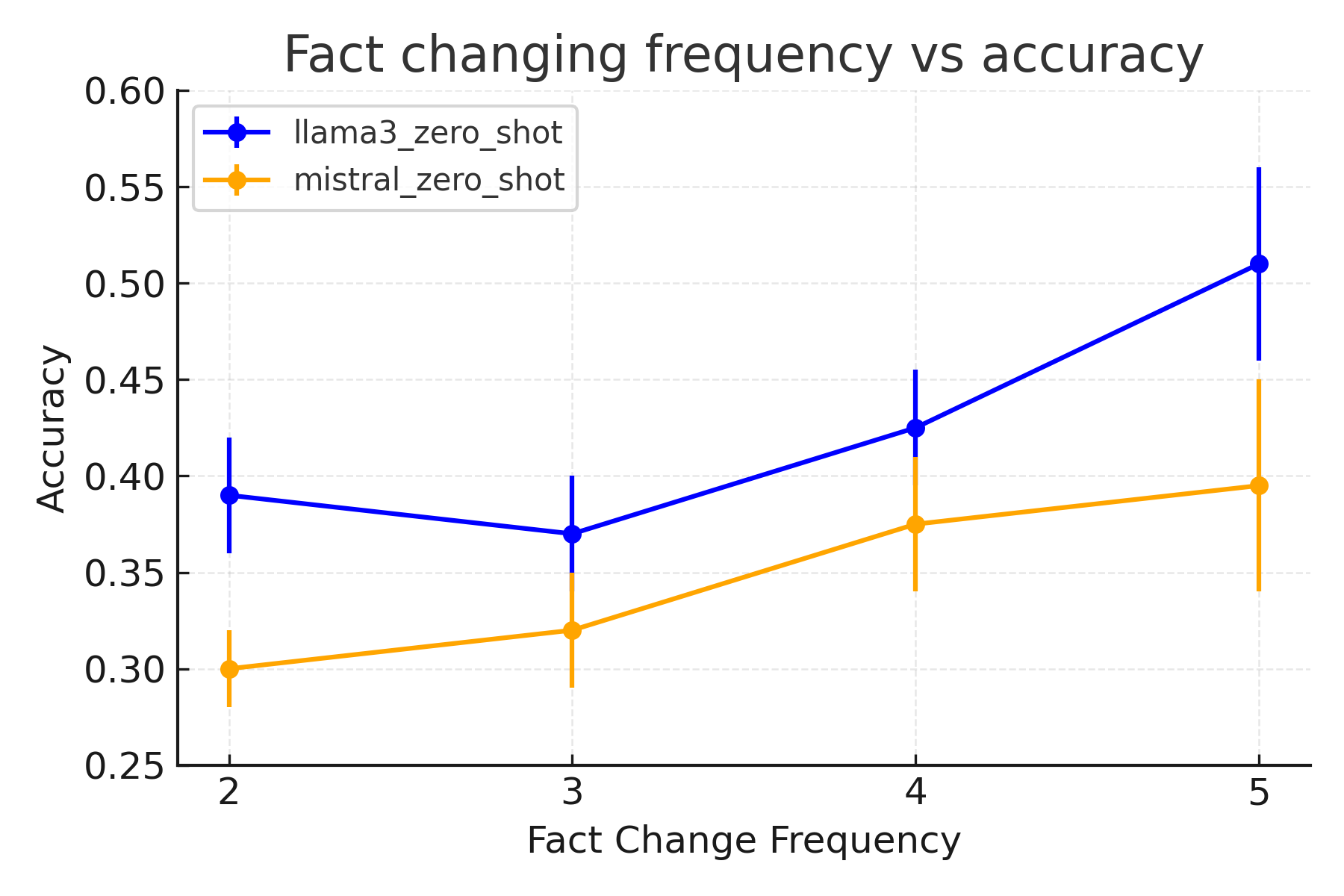}
        \caption{Accuracy plotted against how many times the answer to a question has changed over time.}
        \label{fig:fact-change-accuracy}
    \end{minipage}%
    \hfill
    \begin{minipage}[t]{0.5\textwidth}
        \centering
        \includegraphics[width=\linewidth]{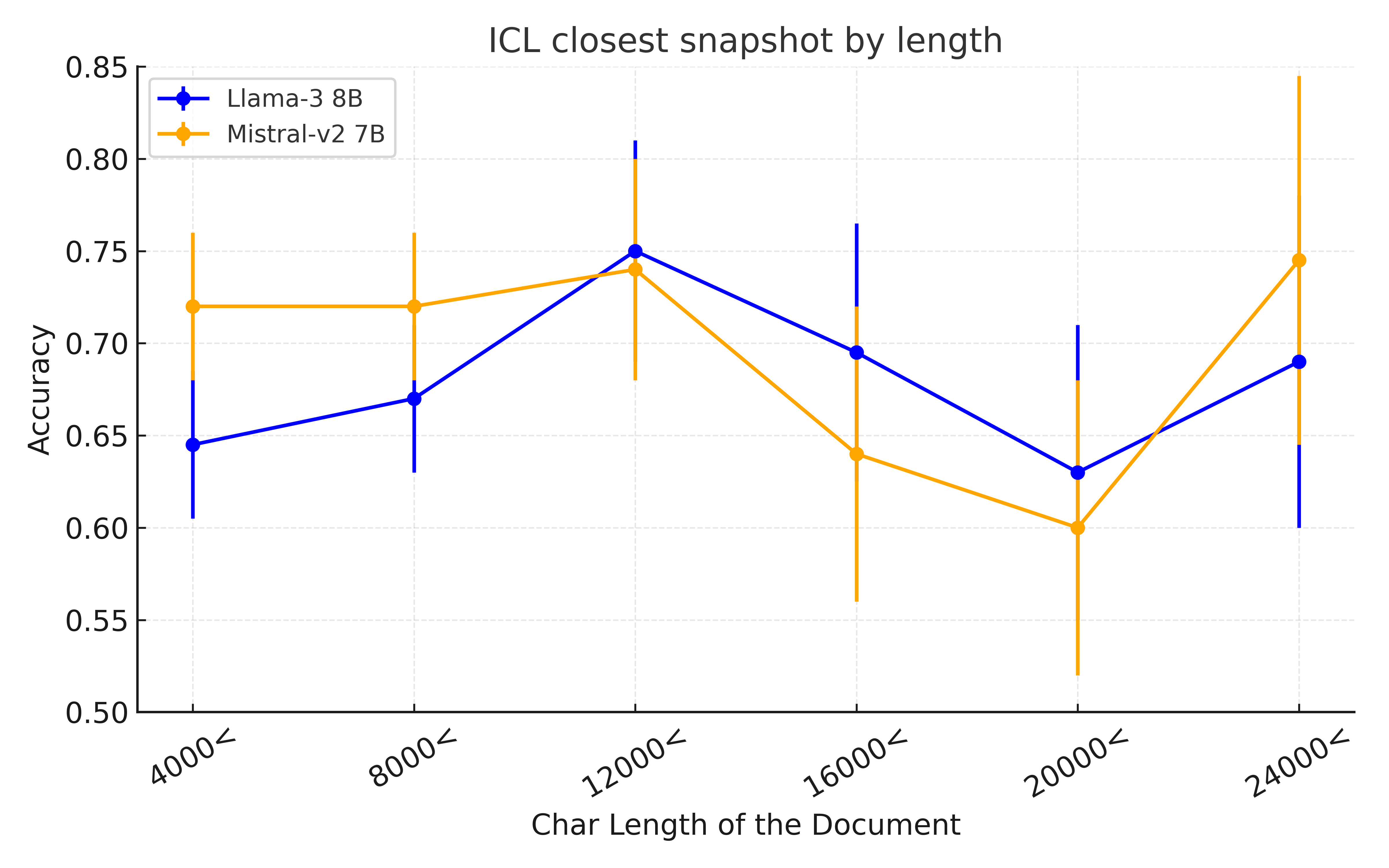}
        \caption{Accuracy plotted against the document length, where we do not observe a consistent pattern.}
        \label{fig:char-length-accuracy}
    \end{minipage}
\end{figure}

We now highlight observations specific to individual experimental setups:

\begin{itemize}
    \item We examined \textbf{zero-shot performance across different fact change frequencies}, as shown in Figure~\ref{fig:fact-change-accuracy}. Interestingly, zero-shot accuracy tends to increase with the frequency of fact changes, that is, with how often the answer to a given question has changed over time. We hypothesize that such frequently updated pages are more likely to have appeared in the LLMs' training data, leading to improved performance without external context.
    \item We also analyzed \textbf{the effect of reference document length} on the ICL setup, which is known to be sensitive to input length. As illustrated in Figure~\ref{fig:char-length-accuracy}, no consistent trend emerged. While accuracy fluctuates across document lengths, these variations do not appear to correlate strongly with document size.
\end{itemize}

In summary, our findings suggest that:
\textit{(i)} temporal context introduces nontrivial challenges, where models can fail to answer correctly even when they succeed without external information, and \textit{(ii)} \textbf{knowledge organization} yields the best performance, followed by RAG, highlighting that distilling complex and temporally conflicting information into structured, concise representations substantially improves accuracy.

\subsection{Unified Clark experiments}

The Unified Clark experiment is designed to evaluate a different capability than the Temporal Wiki benchmark. Specifically, it tests a model's ability to identify all time-dependent instances of a fact within a densely accumulated reference text. Since the input documents often exceed 8,000 tokens, we restrict this experiment to models with extended context length. Additionally, we include GPT-4-o Mini in this setting for additional comparisons. To ensure that all answers are grounded in the provided documents, we disable access to parametric memory by default in this experiment. In other words, models must rely solely on the reference text for accurate responses.

Results are reported in Table~\ref{tab:clark}. As in the Temporal Wiki setting, the \textbf{knowledge organization} approach outperforms all baselines, followed by RAG and then ICL. However, the performance gap is more pronounced here, which we attribute to the greater difficulty of the task: Models must generate complete sets of time-sensitive answers rather than point predictions. Finally, we omit Llama 3 8B results under ICL, as its limited context window cannot accommodate the full reference documents.

\begin{table}[!ht]
    \centering
    \caption{Results on Unified Clark dataset.}
    \vspace{.5cm}
    \begin{tabular}{l|c|c|c}
        \toprule
        \textbf{Model} & \textbf{ICL} & \textbf{RAG} & \textbf{KO} \\
        \midrule
        Gpt4-o Mini & 0.47 & 0.48 & \textbf{0.69} \\
        \midrule
        Llama 3.1 70B & 0.43 & 0.68 & \textbf{0.76} \\
        \midrule
        Qwen 3.1 70B & 0.39 & 0.43 & \textbf{0.71} \\
        \midrule
        Llama-3 8B &  N/A & 0.39 & \textbf{0.69} \\
        \bottomrule
    \end{tabular}
\label{tab:clark}
\end{table}

\section{Discussion}
In this work, we explored how LLMs perform on question answering tasks on temporally evolving knowledge. We introduced two new benchmarks, Temporal Wiki and Unified Clark, to assess model robustness in settings where facts accumulate, shift, or conflict over time. Our results show that while both in-context learning (ICL) and retrieval-augmented generation (RAG) struggle with outdated or conflicting information, RAG consistently outperforms ICL by filtering out irrelevant context. However, both approaches remain vulnerable to temporal noise and may underperform compared to zero-shot baselines. Our proposed knowledge organization framework addresses these challenges by incrementally structuring and indexing facts, enabling more reliable reasoning over dynamic information. These findings highlight the importance of integrating structured temporal memory into future LLM systems.

\paragraph{Acknowledgements} Çağatay Yıldız is a member of the Machine Learning Cluster of Excellence, funded by the Deutsche Forschungsgemeinschaft (DFG, German Research Foundation) under Germany’s Excellence Strategy – EXC number 2064/1 – Project number 390727645. This research utilized compute resources at the Tübingen Machine Learning Cloud, DFG FKZ INST 37/1057-1 FUGG.

\bibliography{bibliography}
\bibliographystyle{iclr2025}

\newpage
\appendix
\section{Appendix}
You may include other additional sections here.

\section*{Temporal Wiki Data Format}

The dataset is structured in JSON format. Each entry in the dataset represents a specific event and contains historical data related to that event at different points in time. The primary structure is a JSON object with two main keys: \texttt{event\_id} and \texttt{incidents}.

\begin{description}
    \item[\texttt{event\_id}:] An integer that serves as a unique identifier for the event.

    \item[\texttt{incidents}:] A JSON object that contains nested objects, where each key is a four-digit string representing a year. Each year-specific object provides a snapshot of the event-related information for that particular year and contains the following fields:
    \begin{description}
        \item[\texttt{q\_year}:] An integer representing the year the question was formulated.
        \item[\texttt{map\_year}:] An integer indicating the year to which the information in the dump pertains, which is used to find the correct answer.
        \item[\texttt{question}:] A string containing the natural language question about the event in that specific year.
        \item[\texttt{answer}:] A list of JSON objects, where each object represents a correct answer. Each answer object contains:
        \begin{itemize}
            \item \texttt{name}: A string with the name of the entity.
            \item \texttt{wikidata\_id}: A string representing the unique Wikidata identifier for the entity.
        \end{itemize}
        \item[\texttt{dump}:] A JSON object containing the raw data retrieved for that year. This includes:
        \begin{itemize}
            \item \texttt{url}: The URL of the Wikipedia page from which the data was sourced.
            \item \texttt{body\_par}: A string containing the main body text of the Wikipedia article.
            \item \texttt{infobox}: A JSON object representing the key-value pairs extracted from the article's infobox.
        \end{itemize}
        \item[\texttt{ans\_comp}, \texttt{llm\_resp}:] These fields are reserved for future use and are currently set to \texttt{null}.
    \end{description}
\end{description}

An example of the data structure is shown below:
\begin{Verbatim}[frame=single,fontsize=\small]
{
    "event_id": 6,
    "incidents": {
        "2010": {
            "q_year": 2010,
            "map_year": 2011,
            "question": "Question: Which team did Luka Modrić play for in 2010?",
            "answer": [
                {
                    "name": "Tottenham Hotspur F.C.",
                    "wikidata_id": "Q18741"
                }
            ],
            "dump": {
                "url": "https://en.wikipedia.org/w/index.php?title=Luka_Modrić",
                "body_par": "Luka Modrić (pronounced , 
                born 9 September 1985 in Zadar)...",
                "infobox": {
                    "Full name": "Luka Modrić[1]",
                    "2008–": "Tottenham Hotspur",
                    ...
                }
            },
            "ans_comp": null,
            "llm_resp": null
        },
        "2013": {
            ...
        }
    }
}
\end{Verbatim}
\end{document}